\documentclass[runningheads]{llncs}

\usepackage[final,year=2024,ID=3272]{eccv}

\usepackage{eccvabbrv}

\usepackage{graphicx}
\usepackage{booktabs}

\usepackage[accsupp]{axessibility}  

\usepackage[pagebackref,breaklinks,colorlinks,citecolor=eccvblue]{hyperref}

\usepackage{orcidlink}

\usepackage{footnote}

\def\eg{\textit{e.g}\onedot} 
\def\ie{\textit{i.e}\onedot}

\makeatother

\newcommand{\pub}[1]{\color{gray}{\tiny{[{#1}]}}}

\usepackage{tabulary}
\newcolumntype{y}[1]{>{\raggedright\arraybackslash}p{#1pt}}
\newcolumntype{z}[1]{>{\raggedleft\arraybackslash}p{#1pt}}

\newcommand{\tablestyle}[2]{\setlength{\tabcolsep}{#1}\renewcommand{\arraystretch}{#2}\centering\footnotesize}

\newcommand{\dataset}{NVI\xspace}
\newcommand{\task}{NVI-DET\xspace}
\newcommand{\hoitask}{HOI-DET\xspace}
\newcommand{\model}{NVI-DEHR\xspace}
\newcommand{\nonverbal}{nonverbal\xspace}

\makeatletter
\newcommand{\thickhline}{%
    \noalign {\ifnum 0=`}\fi \hrule height 0.8pt
    \futurelet \reserved@a \@xhline
}
\makeatother

\let\svthefootnote\thefootnote

\definecolor{bblue}{rgb}{0,150,230}
\definecolor{mygray}{gray}{.9}
\definecolor{myy}{RGB}{126,95,0}
\definecolor{ggray}{RGB}{127,127,127}
\definecolor{mygreen}{RGB}{93,174,86}
\definecolor{myred}{RGB}{255,0,0}
\definecolor{myblue}{RGB}{0,114,188}
\definecolor{darkgreen}{rgb}{0.0, 0.5, 0.0}
\definecolor{demphcolor}{RGB}{100,100,100}
\usepackage{colortbl}
\usepackage{xcolor}
\usepackage{multirow}
\usepackage{makecell}
\usepackage{enumitem}
\usepackage{bm}
\usepackage{threeparttable}

\begin{document}

\title{Nonverbal Interaction Detection}

\titlerunning{Nonverbal Interaction Detection}

\author{Jianan Wei\inst{1*} \and
Tianfei Zhou\inst{2*} \and
Yi Yang\inst{1} \and
Wenguan Wang\inst{1\dag}}

\authorrunning{J.~Wei et al.}

\institute{\small{$^1$Zhejiang University $^2$Beijing Institute of Technology}
\\
\small\url{https://github.com/weijianan1/NVI}}

\maketitle

\begin{abstract}
This work addresses a new challenge of understanding human nonverbal interaction in social contexts. Nonverbal signals pervade virtually every communicative act. Our gestures, facial expressions, postures, gaze, even physical appearance all convey messages, without anything being said. Despite their critical role in social life, nonverbal signals receive very limited attention as compared to the linguistic counterparts, and existing solutions typically examine nonverbal cues in isolation. Our study marks the first systematic effort to enhance the interpretation of multifaceted nonverbal signals. \textbf{First}, we contribute a novel large-scale dataset, called \dataset, which is meticulously annotated to include bounding boxes for humans and corresponding social groups, along with 22 atomic-level nonverbal behaviors under five broad interaction types. \textbf{Second}, we establish a new task \task for nonverbal interaction detection, which is formalized as identifying  triplets in the form  $\langle \texttt{individual}, \texttt{group}, \texttt{interaction} \rangle$ from images. \textbf{Third},  we propose a nonverbal interaction detection hypergraph (\model), a new approach that explicitly models high-order nonverbal interactions using hypergraphs. Central to the model is a dual multi-scale hypergraph that adeptly addresses individual-to-individual and group-to-group correlations across varying scales, facilitating interactional feature learning and eventually improving interaction prediction.   Extensive experiments on \dataset  show that \model improves various baselines significantly in  \task. It also exhibits leading performance on HOI-DET,  confirming its versatility in supporting related tasks and  strong generalization ability. We hope that our study will offer the community new avenues to explore nonverbal signals in more depth.
 \keywords{Nonverbal Interaction \and Hypergraph \and Social Intelligence}
\end{abstract}

\renewcommand{\thefootnote}{}
\footnotetext[1]{*\hspace{0.2em}The first two authors contribute equally to this work.}
\footnotetext[2]{\dag\hspace{0.2em}Corresponding author: Wenguan Wang.}
\renewcommand{\thefootnote}{\svthefootnote}

\section{Introduction}
\label{sec:intro}

\begin{quote}
	\it\small
	Nonverbal  $\cdots$ is an elaborate code written nowhere, known by no one, and understood by all \cite{sapir1949unconscious}. \\
	\mbox{}\hfill -- Edward Sapir (1884 -- 1939)
\end{quote}

\begin{figure}[t]
 	\begin{center}
 	\includegraphics[width=1.0\columnwidth]{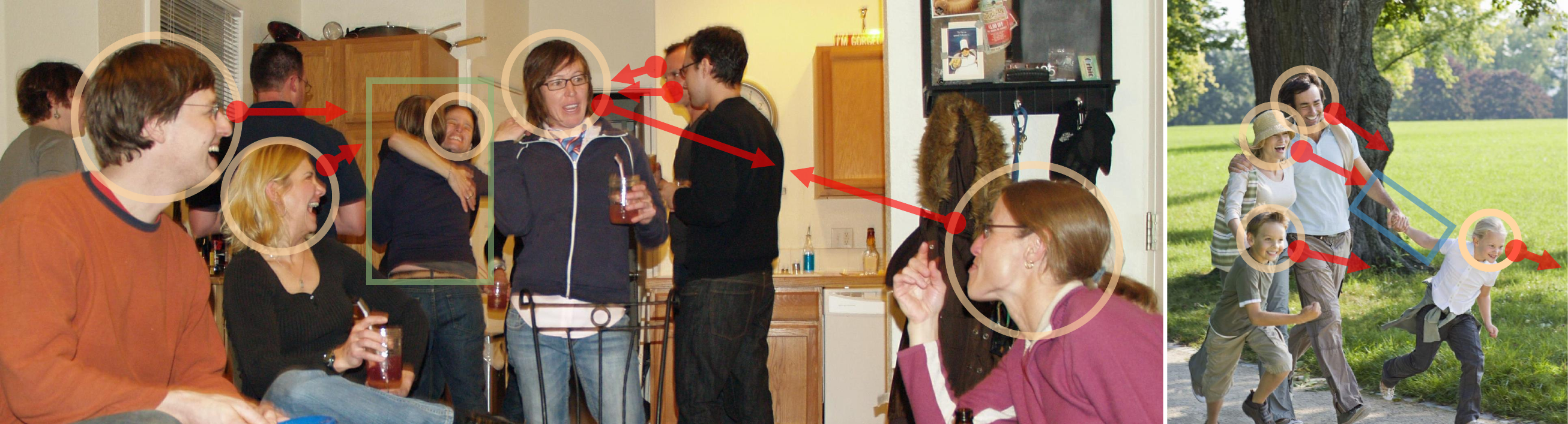}
 	\end{center}
 	\captionsetup{font=small}
 	\caption{\small\textit{\textbf{Can you read these humans?}} Nonverbal interaction (\eg, gaze, gesture) forms the cornerstone of our social life, and serves as the basis of our social intelligence.}
 	\label{fig:motivation}
 \end{figure}

\textit{Nonverbal} behavior allows  us to express intention, attention,  and emotion in our social life. It includes all communicative acts except words,  ranging from facial expression and gaze to bodily contact and physical appearance. 
By analyzing these cues, we gain valuable insights into people's thoughts, feelings, and purposes, even when they are not explicitly stated (Fig. \ref{fig:motivation}).
Though often conducted outside conscious awareness, nonverbal behavior is omnipresent, accounting for nearly two-thirds of all human social interactions \cite{philpott1983relative,pentland2010honest,mehrabian1967inference}. Picture us entering a busy elevator on a Monday morning; while people inside may not seek verbal exchanges, they likely encourage our entry through nonverbal body adjustments.

This primordial form of communication, as widely studied in  cognitive science and psychology \cite{smith2018developing,moore1994social},  is deeply ingrained in human nature. From early infancy, human babies pay extra attention to caregivers' behavior and subsequently learn to convey their needs via nonverbal means (\eg, point, gaze following) long before they speak. 
Endowing machines with the ability to perceive and use non-linguistic interactions is thus an essential step towards giving computers social intelligence \cite{albrecht2006social},  improving human-computer interaction \cite{admoni2014data,andrist2014conversational},  and making them more accessible to people with verbal communication challenges \cite{uluer2023experience}.

Despite its academic value and practical significance, nonverbal interaction analysis has yet to receive sufficient attention in AI research, especially when compared with the immense progress made in  linguistic analysis domains, \ie, NLP and speech recognition.  In computer vision, past efforts  focus on interpreting  nonverbal signals separately, starting from facial expressions \cite{mollahosseini2017affectnet,zhao2011facial}, to gestures  \cite{zhang2018egogesture,narayana2018gesture}, gaze \cite{funes2014eyediap,kellnhofer2019gaze360,fan2019understanding}, and postures \cite{li2010action,xia2012view}, using specialized benchmarks (\eg, RAF-DB \cite{li2017reliable}, VACATION \cite{fan2019understanding}, EgoGesture \cite{zhang2018egogesture}). However, human social interactions are typically characterized by the initiation of concurrent behaviors. For instance, gaze aversion is always accompanied by angry/sad expressions and arm-crossing which indicates a defensive or wary mentality. Similarly, pointing gesture usually triggers gaze-following in others. The specialized datasets fail to capture this complexity nature, and existing systems developed on them are therefore incapable of interpreting them completely \cite{aviezer2012body,meeren2005rapid}.

In this work, we present  \textit{the first} systematical study of  nonverbal interaction in daily social situations. The goal is ambitious -- make machines ``freely'' communicate with humans in a non-linguistic manner. Our solution hinges on  contributions in three fundamental  pillars: \textbf{data}, \textbf{task} and \textbf{model}.

\noindent\textit{\textbf{Data (\S\ref{sec:data}).}}
We introduce \dataset, the first large-scale dataset tailored for \textit{generic} nonverbal interaction understanding. It consists of  {13,711}  images accompanied with high-quality manual annotations of more than  {49K} humans involving in {72K}  social interactions. The image collection covers a wide range of social events such as coffee break in a conference or picnic in a park.  Following previous efforts and established terminologies in social psychology \cite{knapp2013nonverbal,burgoon2021nonverbal,richmond2008nonverbal}, we consider 22  atomic-level  interactions (16 individual- and 6 group-wise), that span across five broad interaction types, \ie, \textit{gaze}, \textit{touch}, \textit{facial expression}, \textit{gesture}, and \textit{posture}. 
\dataset distinguishes itself from all prior datasets in annotation richness of interactions (see Table \ref{table:data}), and opens the door to study the true effect of multifaceted nonverbal signals in human social life.

\noindent\textit{\textbf{Task (\S\ref{sec:task}).}} Upon building \dataset, we define a new task \task to steer the development of AI models towards generic nonverbal interaction understanding. Current efforts in the field is severely diverged -- interactions are described in different ways and addressed in inconsistent granularity, \eg, individual  facial expressions, pair-wise human-object interactions, or group-wise gaze communication. In contrast, we propose that all nonverbal signals, regardless of their broad type and the number of participants, can be distilled into a combination of ``individual behavior'' -- an individual act which nevertheless is influenced by others, and ``collective behavior'' taken together by a group of individuals. With this insight, \task is formulated to localize each individual and the social group it belongs to, meanwhile identifying the category of the interaction. This can be  encapsulated into a triplet $\langle \texttt{individual}, \texttt{group}, \texttt{interaction} \rangle$. This formulation finds relevance to a well-established, yet different task -- human-object interaction detection (HOI-DET) \cite{gupta2015visual}. Unlike HOI-DET that mainly focuses on recognizing actions between human-object pairs, \task  aspires to comprehensively interpret the full spectrum of communactive nonverbal signals, whose patterns are generally more subtle, ambiguous, and  involving multiple persons.

\noindent\textit{\textbf{Model (\S\ref{sec:model}).}}
Clearly, \task is a  structured task that demands a comprehensive modeling of interaction relationships among individuals. To this end, we propose a novel solution named nonverbal interaction detection hypergraph or \model. It is grounded in a hypergraph structure \cite{feng2019hypergraph,zhou2006learning,gao2022hgnn+} that is in nature flexible in high-order relation modeling. Concretely, \model first detects human individuals and social groups following DETR \cite{carion2020end}. It then constructs two distinct multi-scale hypergraphs: one with human individuals as vertices and the other with social groups as vertices. This dual multi-scale hypergraph structure enables deep exploration of both individual-individual and group-group relationships at varying levels of granularity.  Through hypergraph learning, \model obtains  enriched feature representations for individuals and social groups. Finally, \model utilizes these updated features to categorize nonverbal interactions for all individual-group pairs. \model represents a seminal approach for generic nonverbal interaction understanding. We believe that it lays a solid foundation for future research in this fast-evolving domain.

\noindent\textit{\textbf{Experiments (\S\ref{sec:exp}).}}
Our solution is verified through extensive experiments. First, we benchmark the \dataset dataset by modifying a set of HOI-DET models for \task, and offer in-depth discussions on the newly introduced dataset and task. Second, we find that our model yields the best performance in \task, improving the adapted baselines by solid margins. Third, the model exhibits strong  generalization capability, as evidenced by its remarkable performance on two standard HOI-DET benchmarks, \ie, V-COCO \cite{gupta2015visual} and HICO-DET \cite{chao2018learning}.

\begin{table*}[t]
	\centering
	\small
	\captionsetup{font=small}
	\caption{\small\textbf{Comparison of \dataset to related datasets}. Unlike existing datasets that specialize in one particular interaction type, \dataset offers extensive annotations across five common  types,  promoting a unified understanding of nonverbal social interactions.}
	\resizebox{\textwidth}{!}{
		\tablestyle{1pt}{1.3}
		\begin{tabular}{r|r|r|c|c|l|c}
			\hline\thickhline
			\rowcolor{mygray}
			
			& & & & \multicolumn{2}{c|}{Nonverbal Interaction Taxonomy} &  \\ \cline{5-6}
			\rowcolor{mygray}
			\multirow{-2}{*}{Dataset} & \multirow{-2}{*}{Year} & \multicolumn{1}{c|}{\multirow{-2}{*}{Scale}} & \multirow{-2}{*}{Condition}  &  {\# Atomic Class} & \multicolumn{1}{c|}{\# Broad Class} & \multirow{-2}{*}{Task Goal} \\ \hline\hline
			
			MSR-Action3D\!\cite{li2010action}  & 2010
			& 23797 images &controlled& 20& 1 (posture) &  {Action cls.} \\
			UTKinect-Action\!\cite{xia2012view}  & 2012
			& 6220 images &controlled& 10& 1 (posture) &  {Action cls.} \\ \hline
			
			GroupDiscovery\!\cite{li2017reliable} & 2014 
			& 1,176 images & wild & 7 &  1 (posture) & \makecell{Posture cls./det.} \\
			RAF-DB\!\cite{li2017reliable} & 2017
			& 29,672 images & wild & 18 &  1 (facial expression) & \makecell{Facial expression cls.} \\
			Affectnet\!\cite{mollahosseini2017affectnet} & 2017
			& 450,000 images &wild& 8 & 1 (facial expression)& \makecell{Facial expression cls.} \\ \hline
			
			EgoGesture\!\cite{zhang2018egogesture} & 2018
			&  2,081 videos &controlled& 83& 1 (gesture) & Gesture cls./det.\\
			
			LD-ConGR\!\cite{liu2022ld} & 2022
			&  542 videos &controlled& 10& 1 (gesture) & Gesture cls./det./seg.\\ 	\hline
			
			VideoCoAtt\!\cite{fan2018inferring} & 2018
			& 380 videos & wild & 1 & 1 (gaze) & Shared-gaze det. \\
			VACATION\!\cite{fan2019understanding} & 2019
			& 300 videos &wild& 4 & 1 (gaze) & Gaze det.  \\ \hline

			PANDA\!\cite{wang2020panda} & 2020 
			& 600 images & wild & 10 & \makecell{4 (gaze, touch, facial exp-\\ression, posture)} & Human det./trk. \\ \hline \hline

			\textbf{\dataset (Ours)} & 2023& 13,711 images &wild& 22 & \makecell{5 (gaze, touch, facial exp-\\ression, gesture, posture)} & \makecell{Generic nonverbal \\interaction det.} \\
			
			\hline
		\end{tabular}
	}
	\label{table:data}
\end{table*}

\section{Related Work}\label{sec:related}

\noindent\textbf{Nonverbal Interaction Understanding.} Nonverbal communication is a highly efficient and pervasive means for interpersonal exchange. It is integral to our social intelligence \cite{bliege2018social}, which has been argued to be indispensable and  perhaps the most important for success in life \cite{albrecht2006social}. When it comes to computers, however, they are socially ignorant. This gap has led to the emergence of social signal processing (SSP)  \cite{vinciarelli2009social}  that aims at providing computers with the ability to sense and understand human social signals.  Analyzing facial expressions has ever been a core focus in this area \cite{lucey2010extended,zhao2011facial,wang2022ferv39k}, which has achieved tremendous progress. Many other computational methods to automatically detect social cues from images and videos have also been proposed, including recognizing gestures  \cite{zhang2018egogesture,narayana2018gesture}, detecting eye movements \cite{kellnhofer2019gaze360,cheng2020gaze,cheng2020coarse}, inferring emotions  from body posture \cite{schindler2008recognizing}, or detecting social saliency \cite{park20123d}. Despite these advancements, they explore social signals in isolation based on highly specialized datasets, some of which are even created in controlled conditions (see Table~\ref{table:data}). This limits them to interpret subtle meanings of social interactions which are typically transmitted via a combination of multiple signals rather than just one at a time.
Though \cite{alameda2015salsa,joo2019towards} attempt to construct a complex system that can accommodate various SSP recognizers to forecast multiple nonverbal social signals, they necessitate substantial engineering endeavors, posing constrains on the development of this field.

\noindent\textbf{Human Interactions with Objects.} Recently, there has been a strong push on  uncovering human actions with  objects, a task known as HOI-DET \cite{chao2018learning,gkioxari2018detecting,qi2018learning,gao2018ican,kim2020uniondet}. While these actions are also nonverbal, they differ from the types of signals we study in this article. 
Most actions within HOI-DET occur with \textit{explicit intentions}, such as the action \texttt{ride bicycle}. Our work, on the other hand, focuses on the signals that display mostly \textit{implicit intentions} but still produce social awareness. In practice, these unconscious signals are often strong indicators of the initiation of human-object interactions \cite{strabala2012learning}. 
Moreover, HOI-DET deals with pair-wise actions, whereas \task allows for varied forms of higher-order interactions, making it a more nuanced and challenging task.

\noindent\textbf{Hypergraph Learning.} In recent years, graph learning has become prevalent for understanding human-centric visual relationships, notably in human parsing \cite{wang2019learning,wang2020hierarchical,wang2021hierarchical,zhou2021cascaded,zhou2021differentiable,zhou2023differentiable,yang2023large} and HOI detection \cite{qi2018learning,wang2020contextual,gao2020drg,ulutan2020vsgnet,wang2024visual}. Concurrently, hypergraphs have also garnered notable attention for its effectiveness in modeling and learning complex data correlation \cite{gao2020hypergraph,zhou2006learning}. A hypergraph generalizes a standard binary graph; it consists of vertices and hyperedges, and each hyperedge can connect an arbitrary number of vertices, rather than  pair-wise connections in standard graphs. Learning on hypergraphs  then turns into the process of  passing information along hyperedges, facilitating message exchanges among complex relational data. Recently, deep hypergraph learning algorithms \cite{feng2019hypergraph,gao2022hgnn+,jiang2019dynamic,bai2021hypergraph,yadati2019hypergcn} have been proposed and applied to solve computer vision tasks, \eg, image classification \cite{wu2020adahgnn}, pose estimation \cite{xu2022adaptive}, and  mesh reconstruction \cite{huang2023reconstructing}. Drawing inspiration from these advances, we design a hypergraph structure  to model high-order nonverbal interactions among individuals and social groups. By integrating it with a Transformer, our model shows immediate performance improvements, highlighting its efficacy in processing complex nonverbal social signals.

\section{\dataset Dataset} \label{sec:data}

\begin{figure*}[t]
	\begin{center}
	\includegraphics[width=\linewidth]{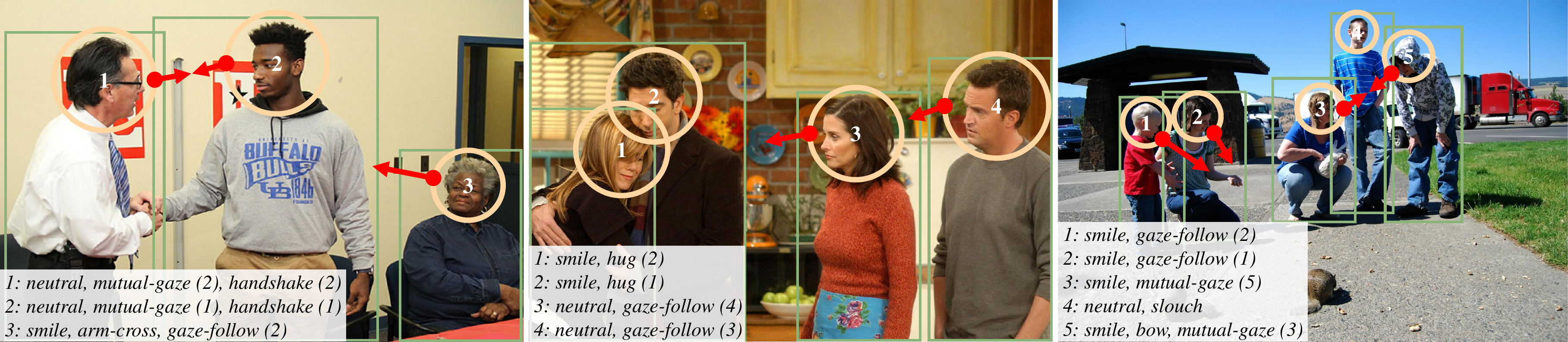}
	\end{center}
	\captionsetup{font=small}
	\caption{\small \textbf{Examples of the \dataset dataset}, showing that our dataset covers rich nonverbal signals in diverse social scenes. Bounding box annotations of individuals are marked by \texttt{green rectangles}. To enhance demonstration and clarity, \texttt{red arrows} and \texttt{numerical identifiers} are incorporated additionally.}
	\label{fig:ex}
\end{figure*}

\dataset is built on PIC 2.0 \cite{liu2021human} that focuses on human-centric relation segmentation. 
PIC 2.0 is labeled solely with segmentation masks for entities (\ie, humans, objects), and  their action/spatial relations (\eg, kicking ball, behind table).  \dataset enriches it by densely labeling social groups (see Fig.~\ref{fig:ex}). To further enhance the quality of \dataset, we exclude images that merely contain a single individual in PIC 2.0. 
Next, we  first define the nonverbal interaction taxonomy used in \dataset (\S\ref{sec:taxonomy}),  then detail the annotation process  (\S\ref{sec:annotation}), and report dataset statistics (\S\ref{sec:statistics}).

\begin{figure*}[t]
	\begin{center}
		\includegraphics[width=\textwidth]{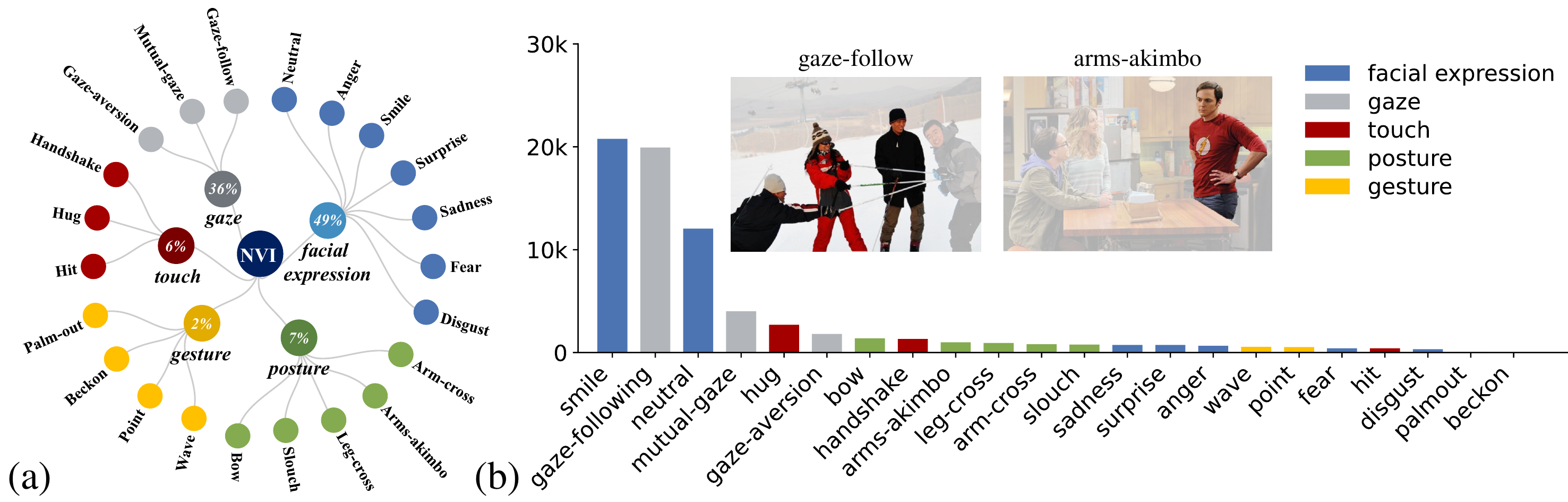}
		\end{center}
	\captionsetup{font=small}
	\caption{\small\textbf{Dataset statistics.} (a) Nonverbal interaction taxonomy (\S\ref{sec:taxonomy}). (b) Distribution of atomic-level nonverbal behaviors (\S\ref{sec:statistics}).}
	\label{fig:stat}
\end{figure*}

\noindent\textbf{Nonverbal Interaction Taxonomy.}\label{sec:taxonomy}
We categorize \nonverbal interactions following a hierarchical taxonomy (see Fig.~\ref{fig:stat}(a)).  We first define five broad interaction types that are recognized as the most important behavioural cues in psychology research \cite{richmond2008nonverbal,hall2019nonverbal,knapp2013nonverbal}. Among them,   two (\ie, \textit{gaze}, \textit{touch}) are group-wise, typically involving multiple individuals, while the other three (\ie, \textit{facial expression}, \textit{gesture}, \textit{posture}) are mainly observed in individuals. Further, each of these broad types is  subdivided into a variety of atomic-level behaviors. The hierarchical taxonomy not only allows our dataset to capture nonverbal signals at multiple levels of abstraction, but also streamlines the annotation process (\S\ref{sec:annotation}).

\begin{itemize}[leftmargin=*]
	\setlength{\itemsep}{0pt}
	\setlength{\parsep}{-2pt}
	\setlength{\parskip}{-0pt}
	\setlength{\leftmargin}{-10pt}
	
	\item\textit{Gaze}, vital for inferring other's intentions \cite{emery2000eyes}, often serves as the first form of communication for interactants. Following  \cite{fan2019understanding}, we consider three specific gaze behaviors in \dataset, \ie, \texttt{gaze-following}, \texttt{mutual-gaze}, and \texttt{gaze-aversion}. 

	\item \textit{Touch}, also known as haptics, is capable of conveying emotions, establishing connections, and  regulating interpersonal dynamics \cite{thayer1986history}. In \dataset, we categorize touch into three atomic behaviors: \texttt{handshake}, \texttt{hug},  and \texttt{hit}.

	\item\textit{Facial expressions} are indicators of people's emotional state, typically conveyed via movements of the lips, eyes, brows, cheeks, and furrows \cite{tian2001recognizing}.  \dataset includes: \texttt{neutral}, \texttt{anger}, \texttt{smile}, \texttt{surprise}, \texttt{sadness}, \texttt{fear}, and \texttt{disgust}.

	\item\textit{Gestures} arise from hand movements, are part of our communicative repertoire from infancy.  \dataset contains four  gestures: \texttt{wave}, \texttt{point}, \texttt{beckon}, and \texttt{palm-out}.
		
	\item\textit{Postures} are  configurations of human body when standing or sitting. In \dataset, we study five classes: \texttt{arm-cross}, \texttt{leg-cross}, \texttt{slouch}, \texttt{arms-akimbo}, and \texttt{bow}.

\end{itemize}

\noindent\textbf{Dataset  Annotation.}\label{sec:annotation}
For each image in \dataset,  annotators are instructed to proceed with the following  steps: \textbf{1)} identify each social group and its broad interaction type $t$; \textbf{2)} determine the specific atomic-level interaction category associated to $t$; \textbf{3)} add bounding boxes for all individuals; \textbf{4)} label the location of the group as the minimal bounding box encompassing all individuals.
Six volunteers are involved in the annotation process. To maintain annotation coverage and accuracy,  annotations of each image are double-checked by a specialist.

\noindent\textbf{Dataset Statistics.}\label{sec:statistics}
\dataset contains 13,711 images in total, and we follow the protocol in PIC 2.0 to split them into
 9,634, 1,418 and 2,659  for \texttt{train}, \texttt{val} and \texttt{test}, respectively. On average, \dataset contains 3.6 human instances, involving in  1.88 gaze, 0.32 touch,  2.60 facial expression, 0.08 gesture, and 0.36 posture per image. The distribution of atomic-level nonverbal behaviors is shown in Fig.~\ref{fig:stat}(b). As seen, \texttt{smile} is the most frequent behavior in \dataset, while \texttt{beckon} rarely occurs.

\section{\task Task}\label{sec:task}

\noindent\textbf{Task Definition.}
As summarized in Table~\ref{table:data}, current research in nonverbal interaction understanding tends to examine different types of interactions independently, each within its own dataset. This hinders AI models from gaining a thorough  understanding of complex human behaviors and poses challenges in properly assessing their true ability in real-world social events. \task is a new task to address this limitation, which encourages models to interpret a full range of nonverbal signals. It has a three-fold objective aimed at: 1) human individual detection, 2) social group detection, and 3) interaction discovery of each individual with their respective group. Formally, accomplishing \task demands the capability to identify all triplets with the form $\langle \texttt{individual}, \texttt{group}, \texttt{interaction} \rangle$.  While seemingly similar to the objective of HOI-DET, \task is much more challenging due to the requirement of distinguishing various heterogeneous nonverbal signals, which are frequently subtle and appear concurrently. 

\noindent\textbf{Metric.} Inspired by \cite{tang2020unbiased,lu2016visual}, we utilize mean Recall@K (mR@K) as our primary evaluation metric, which computes Recall@K (R@K) for each category and then averages all scores:
\begin{equation}\small
mR@K = \frac{1}{|\mathcal{S}^{\tau}|} \sum_{\tau_{iou} \in \mathcal{S}^{\tau}} \frac{1}{C} \sum_{c \in [1 \dots C]} (\sum_{p \in P_c} \mathbb{1}_{\{p \: is \: \text{TP}\}}) / |G_c|,
\end{equation}
where $|\cdot|$ is the cardinality of a set, $\mathcal{S}^{\tau}=\{0.25,0.5,0.75\}$ and $\tau_{iou}$ is the IoU threshold for assigning predicted individuals and social groups to ground truth. And, $C$ is the number of social interaction categories, $P_c$ represents the set of predicted triplets corresponding to the social interaction type $c$, while $G_c$ denotes the set of ground truth triplets associated with the same social interaction type $c$.
This metric continuity reduces barriers to entry. In particular, there are two key reasons why mR@K is employed for our needs: 1) it is more robust to incomplete annotations in \dataset as compared to mean Average Precision (mAP) \cite{lu2016visual}, and 2) it rationally takes into account long-tailed distributions found in \dataset (Fig.~\ref{fig:stat}(b)), since it treats all categories equally. This is a significant advantage over the alternative R@K that exhibits reporting bias \cite{misra2016seeing}. For thorough evaluation,  we adopt mR@25, mR@50, mR@100, along with their average (AR).

\section{\model Model}\label{sec:model}

\begin{figure*}[t]
	\begin{center}
		\includegraphics[width=\textwidth]{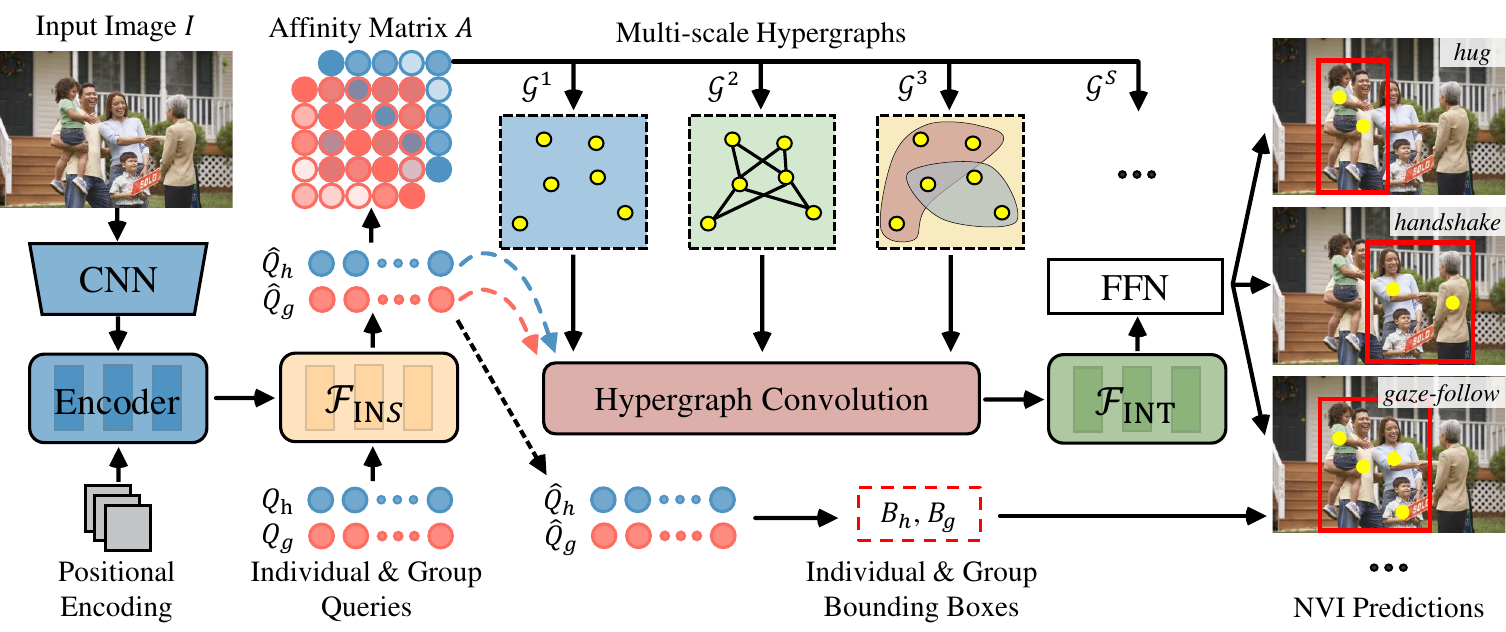}
	\end{center}
	\captionsetup{font=small}
	\caption{\small\textbf{Overall architecture} of the proposed \model model. Given an image, the visual encoder is first applied to extract features, followed by an instance decoder that locates human-object pairs. Next, a dual multi-scale hypergraph is designed to model complex interactions between individuals and social groups via hypergraph convolutions. Lastly, an independent transformer decoder is employed to predict the nonverbal interaction categories for each individual-group pair (\S\ref{sec:model}).}
	\label{fig:ov}
\end{figure*}

\subsection{Preliminary of Hypergraph}\label{sec:preliminary}
A hypergraph can be viewed as a higher-order form of
graph whereby edges can link more than two nodes. Denote $\mathcal{G}\!=\!(\mathcal{V}, \mathcal{E})$ as a hypergraph, where $\mathcal{V}$ is a set of vertices and $\mathcal{E}$ is a set of hyperedges. Each vertex $v\!\in\!\mathcal{V}$ is associated with an initial embedding $\bm{v}$. A hyperedge $e\!\in\!\mathcal{E}$ is a subset of $\mathcal{V}$, indicating the vertices it connects. For convenience, the corresponding hypergraph connectivity structure is usually  represented by a binary incidence matrix $\bm{H}\!\in\!\{0,1\}^{|\mathcal{V}|\!\times\!|\mathcal{E}|}$, where $\bm{H}(v, e)\!=\!1$ if node $v\!\in\!e$, otherwise $\bm{H}(v, e)\!=\!0$. Moreover, we follow  \cite{gao2020hypergraph,feng2019hypergraph} to define the degree of hyperedge $e$ and vertex $v$ as $\delta(e) \!=\! \sum_{v \in \mathcal{V}}\bm{H}(v,e)$ and $d(v) \!=\! \sum_{e \in \mathcal{E}}\bm{H}(v,e)$, respectively.  $\bm{D}_{e} \!\in\! \mathbb{R}^{|\mathcal{E}| \! \times \! |\mathcal{E}|}$ and $\bm{D}_{v} \!\in\! \mathbb{R}^{|\mathcal{V}|\!\times\!|\mathcal{V}|}$  are the diagonal matrices of the hyperedge and  vertex degrees.

\subsection{Nonverbal Interaction Detection Hypergraph}
Fig.~\ref{fig:ov} presents the architecture of our \model. It is built upon the popular encoder-decoder detection structure \cite{carion2020end}, consisting of a shared visual encoder for feature extraction, and two  decoders for  set-based prediction of NVI triplets $\langle \texttt{individual}, \texttt{group}, \texttt{interaction} \rangle$. A unique aspect of the model is that a dual multi-scale hypergraph is introduced to bridge the two decoders. It explicitly models complex interactional contexts among individuals and social groups,  facilitating the learning of high-order cues essential for interaction recognition.

\noindent\textbf{Visual Encoder.} As normal, a visual encoder is adopted to map an input image $I\!\in\!\mathbb{R}^{H_0\!\times\!W_0\!\times\!3}$ into a 3D feature map $\bm{I}\!\in\!\mathbb{R}^{H\!\times\!W\!\times\!D}$. The encoder shares a similar structure as DETR \cite{carion2020end}, consisting of a conventional CNN backbone (\eg, ResNet-50 \cite{he2016deep}) for initial feature extraction, followed by a standard Transformer encoder \cite{carion2020end} to further refine the features by integrating global contextual cues.

\noindent\textbf{Instance Decoder.}
Given $\bm{I}$, our model employs a query-based Transformer decoder to detect human individuals and social groups. The decoder $\mathcal{F}_\text{INS}$ takes two distinct sets of learnable queries as inputs,  \ie,
$\bm{Q}_h\!\in\!\mathbb{R}^{N\!\times\!C}$ and
$\bm{Q}_{g}\!\in\!\mathbb{R}^{N\!\times\!C}$,
and transform them into output embeddings. Subsequently, they are  independently decoded into bounding box coordinates for either  individuals or  groups through a feed forward network (FFN), $\mathcal{F}_\text{FFN}$. Here $\bm{Q}_h$ and $\bm{Q}_g$ serve separately  for the decoding of human individual or social group boxes. $N$ and $C$ denote the number and dimension of these queries.  The entire process can be written as:
\begin{equation}\small\label{eq:instance}
	\begin{aligned}
		\!\!\!\text{query updating:} ~~~~& \hat{\bm{Q}}_h, \hat{\bm{Q}}_g = \mathcal{F}_\text{INS}(\bm{I}, \bm{Q}_h \!+\! \bm{P}, \bm{Q}_g \!+\! \bm{P}), \\
		\!\!\!\text{box prediction:} ~~~~& \bm{B}_h, \bm{B}_g = \mathcal{F}_\text{FFN}( \hat{\bm{Q}_h}, \hat{\bm{Q}_g} ).
	\end{aligned}
\end{equation}
Here we add a learnable position guided embedding $\bm{P}\!\in\!\mathbb{R}^{N\!\times\!C}$ to the queries so as to assign the human query and group query at the same position as a pair \cite{liao2022gen}.
$\hat{\bm{Q}}_h\!=\![\hat{\bm{h}}_1, \hat{\bm{h}}_2, \ldots, \hat{\bm{h}}_N]\!\in\!\mathbb{R}^{N\!\times\!C}$
and
$\hat{\bm{Q}}_g\!=\![\hat{\bm{g}}_1, \hat{\bm{g}}_2, \ldots, \hat{\bm{g}}_N]\!\in\!\mathbb{R}^{N\!\times\!C}$ are the updated queries, while $\bm{B}_h$ and $\bm{B}_g$ denote the box predictions for humans and groups. 

\noindent\textbf{Multi-Scale Hypergraph.} Nonverbal signals are in nature  complicated since they are subtle and often involve multiple  participants. While direct composition of multiple individual features works well for tasks like HOI-DET \cite{liao2022gen,kim2021hotr,chen2021reformulating}, solving \task necessitates a more comprehensive relational understanding of  individuals. To this end,  \model performs multi-scale hypergraph learning  to  explore nonverbal interactions from a multi-granularity perspective.

\textit{Multi-scale hypergraph construction.} 
We utilize two distinct multi-scale hypergraphs, $\mathcal{G}_h$ to model human-human relationships implying that the individuals participate in the same social group, and $\mathcal{G}_g$ to model group-group relationships implying that homologous groups are paired with different individuals.
For clarity, we only explain the construction of $\mathcal{G}_h$, while $\mathcal{G}_g$ follows a same process. Concretely, $\mathcal{G}_h$ is comprised of a set of hypergraphs $\{\mathcal{G}_h^1,\mathcal{G}_h^2,\ldots,\mathcal{G}_h^S\}$, where  $\mathcal{G}_h^s\!=\!(\mathcal{V}_h, \mathcal{E}_h^{s})$ denotes a hypergraph at scale $s$. The vertex set $\mathcal{V}_h$ is consistent across all scales,  and the vertex $v_i\!\in\!\mathcal{V}_h$ represents the $i$-th human query, \ie, we have $|\mathcal{V}_h|\!=\!N$. The hyperedge set $\mathcal{E}_h^s\!=\!\{e_1^s, e_2^s, \ldots, e_{M_s}^s\}$  models group-wise relations with $M_s$ hyperedges, each of which includes $s$ vertices in $\mathcal{V}$. As  in \S\ref{sec:preliminary}, the topology of each $\mathcal{G}_h^s$ is represented by an incidence matrix $\bm{H}^s_h\!\in\!\mathbb{R}^{N\!\times\!M_s}$.

We define the hyperedges based on the distance-based construction strategy \cite{gao2020hypergraph}.  Initially, the embedding of  vertex $v_i$ is set to $\bm{v}_i\!=\!\hat{\bm{h}}_i\!\in\!\mathbb{R}^C$. Then, we compute an affinity matrix $\bm{A}\!\in\!\mathbb{R}^{N\!\times\!N}$ for vertex pairs, wherein each element $\bm{A}_{ij}$ measures the similarity between  $v_i$ and $v_j$:
\begin{equation}\small
	\begin{aligned}
		\bm{A}_{ij} = \bm{v}_{i}^{\top}\bm{v}_{j}/{\| \bm{v}_{i}^{\top} \| \| \bm{v}_{j} \|}.
	\end{aligned}
\end{equation}
Based on the affinity matrix, we form hyperedges at various scales. For the \textit{1st} scale (\ie, $s\!=\!1$),  vertices are independently treated in the graph without any edges and the incidence matrix $\bm{H}_h^0$ is thus an identity matrix. For other scales, each hyperedge $e_i^s$ is formed as a cluster of vertices, which is identified by searching for a $s\!\times\!s$  sub-matrix within $\bm{A}$ that exhibits the highest density:
\begin{equation}\small\label{eq:submatrix}
	e_i^s = \arg\max_{\mathcal{O}\subseteq\mathcal{V}_h} ||\bm{A}_{\mathcal{O},\mathcal{O}}||_{1,1}, ~~~\textit{s.t.}~~ |\mathcal{O}|=s ~~\text{and}~~ v_i\in\mathcal{O}.
\end{equation}
Here $\|\cdot\|_{1,1}$ represents the $L_{1,1}$ matrix norm, \ie,  the summation of  the absolute values of all elements in the matrix. The objective of Eq.~\ref{eq:submatrix} is to identify and connect the most closely related vertices.  The first constraint confines the size of each cluster, while the second ensures the inclusion of $v_i$ in the group. The optimization problem is tackled via a vertex-centric greedy algorithm, which, at each iteration $i$, selects the vertex $v_i$ first and then includes additional $s\!-\!1$ vertices based on their affinity to $v_i$.  In this way, for scales $s\!>\!1$, we have  $M_s\!=\!N$.

\textit{Multi-scale hypergraph learning.} With multi-scale hypergraphs $\mathcal{G}_h$ and $\mathcal{G}_g$, we perform message exchange among vertices  through a sequence of  $L$  hyperedge convolutional layers \cite{feng2019hypergraph}. For each scale $s$, the convolution operation at layer $l\!\in\![L]$ can be formulated as:
{\small
\begin{align}
		\!\!\!\!\bm{V}^{s,(l)}_h &\!\!=\!\! (\bm{D}^s_{h,v})^{\!-\!\frac{1}{2}} \bm{H}^s_h  (\bm{D}^{s}_{h,e})^{\!-1} \bm{H}_h^{s\top} (\bm{D}^{s}_{h,v})^{\!-\!\frac{1}{2}} \bm{V}_h^{s,(l\!-\!1)} \theta_h^{s,(l)}, \label{eq:humanconv}\\
		\!\!\!\!\bm{V}^{s,(l)}_g &\!\!=\!\! (\bm{D}^s_{g,v})^{\!-\!\frac{1}{2}} \bm{H}^s_g  (\bm{D}^{s}_{g,e})^{\!-1} \bm{H}_g^{s\top} (\bm{D}^{s}_{g,v})^{\!-\!\frac{1}{2}} \bm{V}_g^{s,(l\!-\!1)} \theta_g^{s,(l)}, \label{eq:groupconv}
\end{align}}
\!\!where Eq.~\ref{eq:humanconv} and Eq.~\ref{eq:groupconv} are applied for human hypergraph and group hypergraph, respectively. The $\theta_h^{s,(l)}$ and $\theta_g^{s,(l)}$ are learnable parameters of the $l$-th  layer at scale $s$. $\bm{D}_{\cdot,v}^s/\bm{D}_{\cdot,e}^s$ denote diagonal matrices of  vertex and hyperedge degrees for either $\mathcal{G}_h^s$ or $\mathcal{G}_g^s$, which are computed from corresponding incidence matrices (\S\ref{sec:preliminary}).  $\bm{V}_h^{s,(0)\!}/\bm{V}_g^{s,(0)}$ are  matrices with initial vertex embedding in $\mathcal{G}_h^s/\mathcal{G}_g^s$.
After $L$ convolutions, the final embedding for each vertex is obtained by aggregating information across various scales via a multilayer perceptron (MLP):
\begin{equation}\small
	\begin{aligned}
	\bm{F}_h &=  \text{MLP}([\bm{V}_h^{1,(L)}, \bm{V}_h^{2,(L)}, \ldots, \bm{V}_h^{S,(L)}])~~\in\mathbb{R}^{N\times C}, \\
	\bm{F}_g &=  \text{MLP}([\bm{V}_g^{1,(L)}, \bm{V}_g^{2,(L)}, \ldots, \bm{V}_g^{S,(L)}])~~\in\mathbb{R}^{N\times C},
	\end{aligned}
\end{equation}
where $\bm{F}_h$ and $\bm{F}_g$ are the final embedding matrices of all human and group vertices, respectively. $'[\cdot,\cdot]'$ denotes tensor concatenation.

\noindent\textbf{Interaction Decoder.} Last, we leverage an independent query-based Transformer decoder to predict the nonverbal interaction categories for each  individual-group pair. Instead of random query initialization, we propose to  create nonverbal interaction query $\bm{Q}_n\!\in\!\mathbb{R}^{N\!\times\!C}$ dynamically based on high-order features of individuals $\bm{F}_h$ and  groups $\bm{F}_g$:
\begin{equation}\small
	\bm{Q}_n = (\bm{F}_h + \bm{F}_g) / 2.
\end{equation}
Note that  these two types of features  can be directly added, since they are  position-aligned in Eq.~\ref{eq:instance}. Subsequently, the interaction decoder takes the  query $\bm{Q}_n$ alongside image feature $\bm{I}$ as input, and predict  interaction categories as:
\begin{equation}\small\label{eq:interactiondecoder}
	\bm{P} = \mathcal{F}_\text{INT}(\bm{I}, \bm{Q}_n),
\end{equation}
where $\bm{P}$ is the nonverbal interaction predictions for $N$ individual-group pairs.

\subsection{Detailed Network Architecture}
\noindent\textbf{Network Architecture.}
We utilize ResNet-50 \cite{he2016deep} as the CNN backbone in all experiments. Following DETR \cite{carion2020end}, the Transformer encoder  consists of six standard Transformer layers, while both the instance $\mathcal{F}_\text{INS}$ (\textit{cf.} Eq.~\ref{eq:instance}) and interaction $\mathcal{F}_\text{INT}$ (\textit{cf.} Eq.~\ref{eq:interactiondecoder}) Transformer decoders incorporate three layers. By default, we set the number of  queries  $N$ to 64, the number of channels $C$ to 256, the number of hypergraph scales  $S$ to 5, and adopt $L\!=\!2$ hyperedge convolutional layers.

\noindent\textbf{Training Objective.}
We follow \cite{tamura2021qpic,carion2020end,liao2022gen} to perform end-to-end training by assigning a bipartite matching prediction with each groundtruth using the Hungarian algorithm.  The   loss function is:
 $\mathcal{L} \!=\! \lambda_{1} \mathcal{L}_\text{1} \!+\! \lambda_2 \mathcal{L}_\text{GIoU} \!+\! \lambda_3 \mathcal{L}_\text{c}$, consisting of three parts: a $\mathcal{L}_1$ loss and a generalized IoU loss $\mathcal{L}_\text{GIoU}$ to assess localization accuracy, a focal loss $\mathcal{L}_\text{c}$ to evaluate interaction classification. The coefficients are empirically set to: $\lambda_1\!=\!2.5$, $\lambda_2\!=\!1$, $\lambda_3\!=\!2$, in accordance with QPIC \cite{tamura2021qpic}.

\noindent\textbf{Reproducibility.} Our model is implemented using PyTorch and trained on 4 NVIDIA GeForce RTX 3090 GPUs. Testing is carried out on the same machine. 

\section{Experiment}\label{sec:exp}

\subsection{Experiment on \task}\label{sec:expnvi}

\begin{figure*}[t]
	\begin{minipage}{\textwidth}
		\makeatletter\def\@captype{table}\makeatother\captionsetup{font=small}\caption{\small\textbf{NVI-DET results} on \dataset \texttt{val} and \texttt{test} (\S\ref{sec:expnvi}).  \label{tab:hnb}}
		\begin{threeparttable}
			\resizebox{0.98\textwidth}{!}{
				\tablestyle{4pt}{1.1}
				\begin{tabular}{r||c c c c|c c c c}\hline\thickhline
					\rowcolor{mygray}
					& \multicolumn{4}{c|}{\texttt{val}}& \multicolumn{4}{c}{\texttt{test}}\\
					\rowcolor{mygray}
					\multirow{-2}{*}{Method} &  mR@25& mR@50& mR@100& AR& mR@25& mR@50& mR@100& AR \\ \hline\hline		
					$m$-QPIC\!\cite{tamura2021qpic}& \textbf{56.89}& 69.52& 78.36& 68.26& 59.44& 71.46& 80.07& 70.32\\
					$m$-CDN\!\cite{zhang2021mining}& 55.57& 71.06& 78.81& 68.48& 59.01& 72.94& 82.61& 71.52\\ 
					$m$-GEN-VLKT\!\cite{liao2022gen}& 50.59& 70.87& 80.08& 67.18& 56.68& 74.32& 84.18& 71.72\\ \hline 			
					\textbf{\model} (\textbf{Ours})& 54.85& \textbf{73.42}& \textbf{85.33}& \textbf{71.20}& \textbf{59.46}& \textbf{76.01}& \textbf{88.52}& \textbf{74.67}\\ \hline
				\end{tabular}
			}
		\end{threeparttable}
	\end{minipage}
\end{figure*}

\begin{figure*}[t]
	\begin{minipage}{\textwidth}
		\makeatletter\def\@captype{table}\makeatother\captionsetup{font=small}\caption{\small\textbf{Results} for individual- and group-wise interactions on \dataset \texttt{val} (\S\ref{sec:exphoi}).  \label{tab:ig} }
		\begin{threeparttable}
			\resizebox{0.98\columnwidth}{!}{
				\tablestyle{4pt}{1.1}
				\begin{tabular}{r||c c c c|c c c c}\hline\thickhline
					\rowcolor{mygray}
					& \multicolumn{4}{c|}{individual}& \multicolumn{4}{c}{group}\\
					\rowcolor{mygray}						
					\multirow{-2}{*}{Method} & mR@25& mR@50& mR@100& AR& mR@25& mR@50& mR@100& AR \\ \hline\hline		
					$m$-QPIC\!\cite{tamura2021qpic}& \textbf{52.23}& 66.09& 75.98& 64.77& \textbf{69.18}& 78.62& 84.85& 77.55\\
					$m$-CDN\!\cite{zhang2021mining}& 50.67& 68.23& 76.74& 65.21& 68.66& 78.60& 84.34& 77.20\\
					$m$-GEN-VLKT\!\cite{liao2022gen}& 44.98& 68.51& 78.30& 63.93& 67.84& 79.47& 87.12& 78.14\\ \hline
					\textbf{\model} (\textbf{Ours})& 49.37& \textbf{70.04}& \textbf{83.82}& \textbf{67.74}& 69.47& \textbf{82.45}& \textbf{89.35}& \textbf{80.42}\\ \hline
				\end{tabular}
			}
		\end{threeparttable}
	\end{minipage}
\end{figure*}

\textbf{Competitors.} To better benchmark \dataset and verify the proposed model, we adapt three state-of-the-art \hoitask approaches,  \ie, QPIC \cite{tamura2021qpic}, CDN \cite{zhang2021mining}, GEN-VLKT \cite{liao2022gen}, for nonverbal interaction detection, denoted as $m$-QPIC, $m$-CDN, and $m$-GEN-VLKT. 
We modify their interaction prediction head to align with our \task task. For $m$-GEN-VLKT that relies on CLIP, we modify its text prompt to the format of `A photo of a person [nonverbal interaction]'.

\noindent\textbf{Implementation Details.} For fairness, we train all models for 90 epochs. The learning rate is set to be 1e-4 for the initial 60 epochs and decreased to 1e-5 for the remaining epochs. AdamW is used as the optimizer. Common  data augmentation techniques are applied,  including random horizontal flipping, color jittering, and random scaling. Moving on to the procedural details during training, it is important to note that the bounding box of social group for ``individual interaction'' is identical to the individual bounding box. During inference, the ground-truths of triplets that encompass individual interactions do not entail bounding box of social group, akin to body motion categories in V-COCO \cite{gupta2015visual}.

\noindent\textbf{Quantitative Results.}
Table~\ref{tab:hnb} presents the benchmarking results on \dataset \texttt{val} and \texttt{test}. As seen, $m$-QPIC, which is a simple adaptation of DETR from object detection to interaction detection, produces the worst performance among the comparative approaches due to the lack of explicit relational reasoning. $m$-CDN performs much better than $m$-QPIC by introducing an independent interaction decoder to account for interactional relations. Notably, the CLIP-based method $m$-GEN-VLKT performs worse in \task,  revealing that  transferring knowledge from visual language models to \task seems to be more difficult than to \hoitask. \model surpasses all the baselines, reaching \textbf{71.20} and \textbf{74.67} AR on \dataset \texttt{val} and \texttt{test}, respectively.
Furthermore, we explore how the models perform  for individual- and group-wise interactions. Table~\ref{tab:ig} shows our model  consistently outperforms others in both sets across all  metrics except for mR@25.

\begin{figure*}[t]
	\begin{center}
		\includegraphics[width=\linewidth]{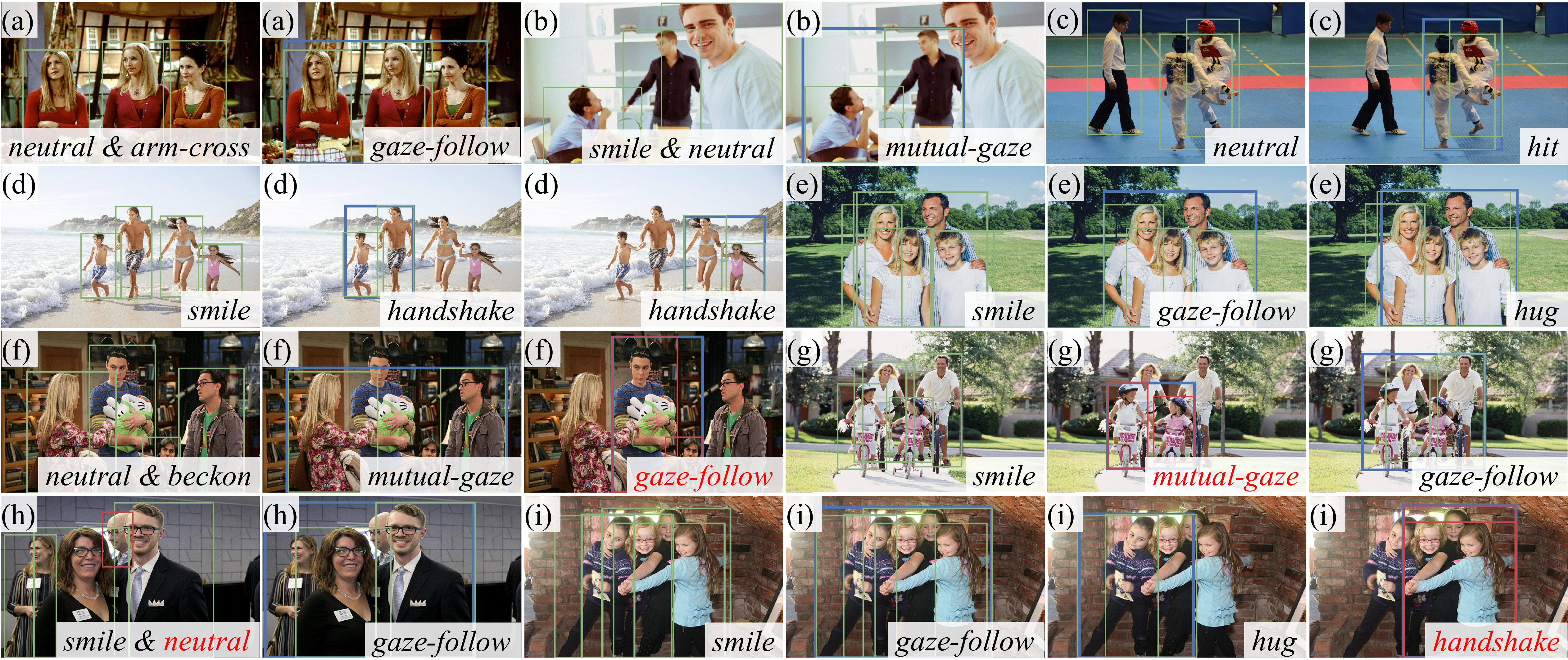}
	\end{center}
	\captionsetup{font=small}
	\caption{\small\textbf{Visualization of \task} results on \dataset \texttt{val}. Predictions are generated with $\{K=30, \tau_{iou}=0.5\}$. Individuals correctly identified are in green bounding boxes, whilst those not detected are in red. Blue bounding boxes are utilized to denote social groups. True positive predictions are in black font, while false positives are in red (\S\ref{sec:expnvi}).}
	\label{fig:visual}
\end{figure*}

\noindent\textbf{Qualitative Results.}
In Fig.~\ref{fig:visual}, we show nonverbal interaction detection results of our model on \dataset \texttt{test}. 
Our model can accurately localize the interactions, and successfully recognize different nonverbal interactions.
For instance, in case (e), it competently forms a gaze-following group from three out of four individuals based on their gaze direction, and also astutely recognizes the man's hugging behavior in the image, although the bounding box for the hugging group is somewhat undersized.
We also present some failure cases (shown in red), which may be due to the ambiguity and subtlety of nonverbal interactions.
In case (g), the mutual gaze of the two little girls could easily be mistaken for gaze aversion, as their gazes do not align perfectly.
Likewise, the handshake behavior in case (i) could be confused with hugging, due to the unusual pose involved.
It is notable that individuals heavily obscured in the image also present a challenge to the \task, as shown in case (h).
These results highlight the challenging aspect of the \task task, meanwhile offering opportunities for future exploration.

\begin{table}[t]
	\centering
	\small
	\caption{\small\textbf{HOI-DET results} on HICO-DET \cite{chao2018learning} \texttt{test} and V-COCO \cite{gupta2015visual} \texttt{test}$_{\!}$ (\S\ref{sec:exphoi}).} \label{tab:hoi}
	\resizebox{\columnwidth}{!}{
		\tablestyle{8pt}{1.05}
		\begin{tabular}{r@{}l@{}|l||c c c|cc}\hline\thickhline
			\rowcolor{mygray}
			& & & \multicolumn{3}{c|}{HICO-DET}&  \multicolumn{2}{c}{V-COCO}\\
			\rowcolor{mygray}
			\multicolumn{2}{c|}{\multirow{-2}{*}{Method}} & \multirow{-2}{*}{Backbone}& Full& Rare& Non-Rare& $AP^{S1}_{role}$ & $AP^{S2}_{role}$ \\ \hline\hline
			InteractNet\!\cite{gkioxari2018detecting}&\pub{CVPR18}& R50-FPN& 9.94& 7.16& 10.77& 40.0& 48.0 \\
			UnionDet\!\cite{kim2020uniondet}&\pub{ECCV20}& R50& 17.58& 11.72& 19.33& 47.5& 56.2\\
			PPDM\!\cite{liao2020ppdm}&\pub{CVPR20}& HG-104& 21.73& 13.78& 24.10& -& -\\
			HOTR\!\cite{kim2021hotr}&\pub{CVPR21}& R50& 23.46& 16.21& 25.60& 55.2& 64.4\\
			QPIC\!\cite{tamura2021qpic}&\pub{CVPR21}& R50& 29.07& 21.85& 31.23& 58.8& 61.0\\
			ODM\!\cite{wang2022chairs}&\pub{ECCV22}& R50-FPN& 31.65& 24.95& 33.65& -& -\\
			UPT\!\cite{zhang2022efficient}&\pub{CVPR22}& R50& 31.66 &25.94 &33.36& 59.0& 64.5\\
			CDN\!\cite{zhang2021mining}&\pub{NeurIPS21}& R50& 31.78& 27.55& 33.05& 62.3& 64.4\\
			Iwin\!\cite{tu2022iwin}&\pub{ECCV22}& R50-FPN& 32.03& 27.62& 34.14& 60.5& -\\
			DOQ\!\cite{qu2022distillation}&\pub{CVPR22}& R50& 33.28& 29.19& 34.50& 63.5& -\\
			GEN-VLK\!\cite{liao2022gen}&\pub{CVPR22}& R50& 33.75& 29.25& 35.10& 62.4& 64.4\\
			HOICLIP\!\cite{ning2023hoiclip}&\pub{CVPR23}&R50& 34.54& 30.71& 35.70& 63.5& 64.8\\ \hline
			\textbf{\model} &(\textbf{Ours})& R50& \textbf{35.30}& \textbf{31.43}& \textbf{36.64}& \textbf{64.1}& \textbf{65.3}\\ \hline
		\end{tabular}
	}
	\captionsetup{font=small}
\end{table}

\begin{table}[t]
	\begin{minipage}[b]{0.49\linewidth}
	\caption{\small\textbf{Analysis of multi-scale hypergraph} on \dataset \texttt{val} (\S\ref{sec:ablationstudy}).} \label{tab:gs}
	\centering
	\resizebox{\columnwidth}{!}{
			\tablestyle{6pt}{1.03}
			\begin{tabular}{c||c c c c}\hline\thickhline
				\rowcolor{mygray}
				$S$& mR@25& mR@50& mR@100& AR\\
				\hline
				\hline
				1& 53.39& 69.81& 81.90& 68.37\\
				2& 53.44& 70.31& 82.62& 68.79\\
				3& 53.52& 70.45& 83.92& 69.30\\
				4& 54.21& 73.34& 84.36& 70.64\\
				5& \textbf{54.85}& \textbf{73.42}& \textbf{85.33}& \textbf{71.20}\\
				6& 54.59& 73.11& 85.24& 70.98\\
				\hline
			\end{tabular}
		}
		\captionsetup{font=small}
	\end{minipage}
	\hspace{2pt}
	\begin{minipage}[b]{0.49\linewidth}
	\caption{\small\textbf{Impact of hyperedge convolutions} on \dataset \texttt{val} (\S\ref{sec:ablationstudy}).} \label{tab:ln}
	\centering
	\resizebox{\columnwidth}{!}{
		\tablestyle{6pt}{1.45}
		\begin{tabular}{c||c c c c}\hline\thickhline
			\rowcolor{mygray}
			$L$ & mR@25& mR@50& mR@100& AR\\
			\hline
			\hline
			0& 53.50& 69.44& 81.71& 68.22\\ 
			1& 53.76& 71.74& 83.61& 69.70\\
			2& \textbf{54.85}& \textbf{73.42}& \textbf{85.33}& \textbf{71.20}\\
			3& 54.36& 72.47& 84.85& 70.56\\
			\hline
		\end{tabular}
	}
	\captionsetup{font=small}
	\end{minipage}
\end{table}

\subsection{Experiment on \hoitask}\label{sec:exphoi}
\textbf{Datasets.}
To fully assess the capability of our model, we evaluate it in the \hoitask using two popular datasets, \ie, V-COCO \cite{gupta2015visual} and HICO-DET \cite{chao2018learning}.

\noindent\textbf{Evaluation Metrics.}
In accordance with conventions \cite{gupta2015visual,gkioxari2018detecting,qi2018learning,zhou2021cascaded}, the mean average precision (mAP) calculated  on HOI triplets is used as the metric.

\noindent\textbf{Implementation Details.}
During training, we use the \hoitask loss in \cite{liao2022gen,ning2023hoiclip} for network optimization.
The model is trained with the same setting used in Section \ref{sec:expnvi}.
Inspired by  \cite{liao2022gen,iftekhar2022look,yuan2022rlip,wang2022learning,qu2022distillation}, our HOI classifier is initialized with the HOI embeddings generated by the text encoder of CLIP, and adopts a variant of the cross-attention module \cite{ning2023hoiclip}  to utilize visual representation from CLIP.
During inference, we first filter out low-confidence predictions and then  apply non-maximum suppression to reduce overlapped predictions, following \cite{liao2022gen,yuan2022rlip,qu2022distillation}.

\noindent\textbf{Quantitative Results.}
Table~\ref{tab:hoi} presents the detection performance of  our model against 12 state-of-the-art \hoitask methods in the two datasets . Our model achieves the best performance in all settings.
Particularly, in the HICO-DET dataset, our model outperforms the best-performing approach \cite{ning2023hoiclip}, with promising gains of  0.76, 0.72, and 0.94 for Full, Rare, and Non-Rare categories, respectively.
In V-COCO, our approach achieves scores of $\textbf{64.1}$ in terms of $AP^{S1}_{role}$   and $\textbf{65.3}$ in terms of $AP^{S2}_{role}$. These results verify the flexibility and generalizability of our model in handling various relational reasoning tasks.

\subsection{Diagnostic Experiment} \label{sec:ablationstudy}

\noindent\textbf{Number of Hypergraph Scales $S$.}
We first examine how the number $S$ of scales  in multi-scale hypergraph construction impacts model performance, as reported in Table~\ref{tab:gs}. 
We observe that the model obtains an AR score of 68.37 at $S\!=\!1$, with all humans and social groups are treated as independent vertices in the hypergraph. As $S$ rises, model performance progressively improves, reaching a best performance of $71.20$ AR at $S\!=\!5$.
It aligns with the research conducted by \cite{fay2000group}, which underscores that the majority of the speech in discussions involving 10 or more participants is produced by only the top 4-5 contributors. 

\noindent\textbf{Number of Hyperedge Convolutions $L$.}
In Table~\ref{tab:ln}, we  examine the effect of the number $L$ of hypergraph convolution layers. Here $L\!=\!0$ represents a variant of our model without hypergraph learning. It yields a score of $68.22$ AR. By introducing more layers, model performance improves as $L$ increases, and tends to stabilize at $L\!=\!2$, with a significant boost to $71.20$. We argue that increasing $L$ leads to noise spreading in the graph, which could impair the final prediction.

\section{Conclusion}
This work makes a substantial step towards automatic interpretation of human nonverbal  behaviors in everyday social environments. We challenge  the conventional paradigm, which isolates social signals for standalone study, by instead examining a variety of common nonverbal signals (\ie, \textit{gaze}, \textit{facial expression}, \textit{gesture}, \textit{posture}, \textit{touch}) collectively. To open this avenue,  we create a richly annotated  dataset \dataset, formalize the nonverbal interaction detection task \task, and devise a baseline model \model based on hypergraph learning. \model model achieves impressive performance in two interaction detection tasks, \ie, \task and \hoitask. 
Despite this, our experiments reveal that \task is considerably complex, regarding pluralistic relation among individuals, and we are now far from tackling this problem. 
We hope that our study will serve as valuable resources to foster more extensive exploration in this field.

%
%
\bibliographystyle{splncs04}
\bibliography{nvi}

\newpage
\onecolumn
  \null
  \vskip .375in
  \begin{center}
    {\Large \bfseries Supplemental Material \par}
\end{center}

\renewcommand\thesection{\Alph{section}}
\renewcommand{\thetable}{S\arabic{table}}
\renewcommand{\thefigure}{S\arabic{figure}}

\setcounter{section}{0}
\setcounter{figure}{0}
\setcounter{table}{0}

This document provides more details to supplement our main manuscript.
We first give additional analyses about \dataset in \S\ref{sec:stat} and present more implementation details on HOI-DET in \S\ref{sec:hoi}.
Subsequently, additional quantitative results of our \model are summarized in \S\ref{sec:quant}.
Finally, we delve into an in-depth discussion about social impact, potential limitations and future directions in \S\ref{sec:disc}.

\section{Additional Dataset Analysis} \label{sec:stat}
\begin{figure}[h]
	\begin{center}
		\includegraphics[width=0.8\textwidth]{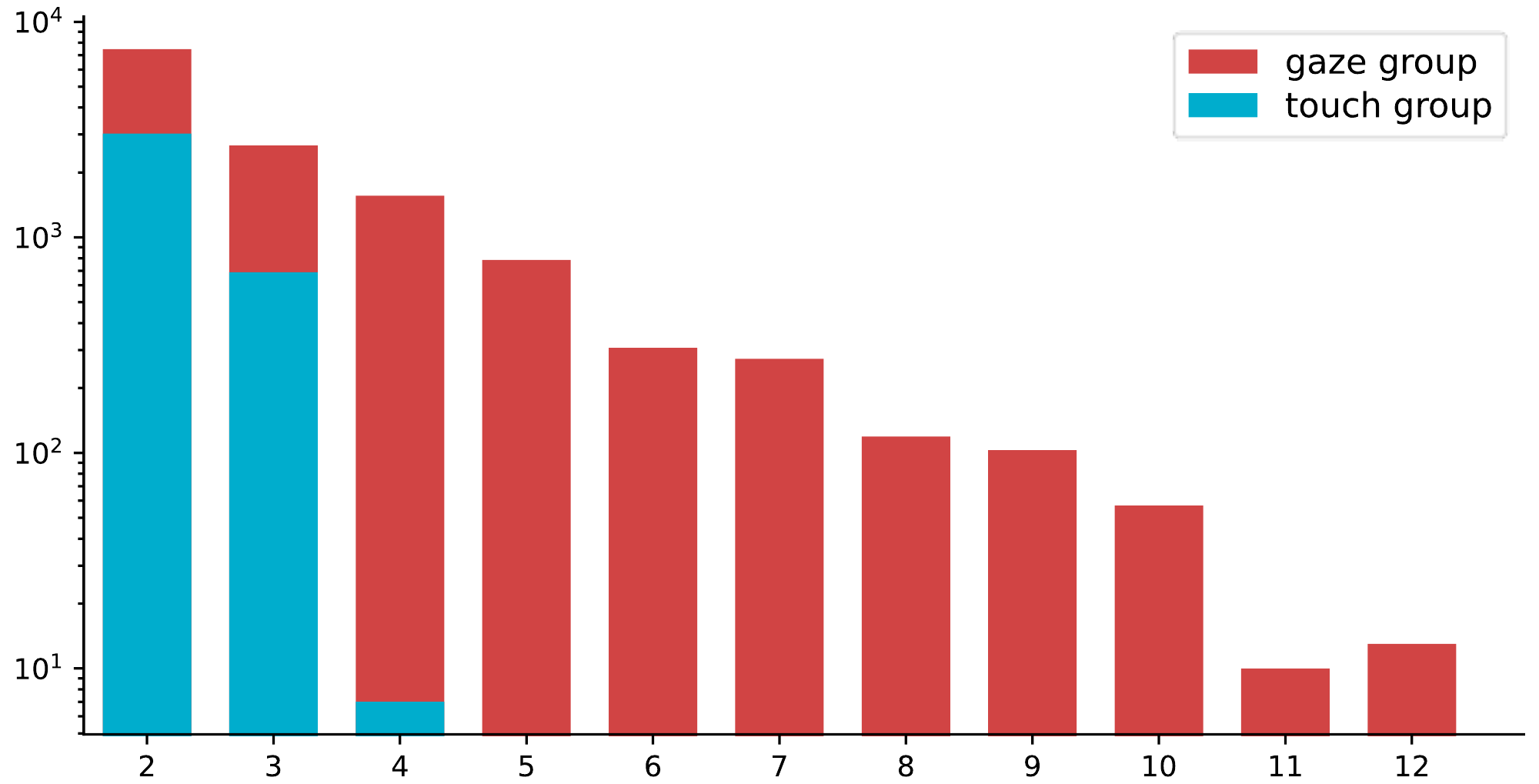}
	\end{center}
	\captionsetup{font=small}
	\caption{\small\textbf{Group size} in group-wise interactions (\S\ref{sec:stat}).}
	\label{fig:dis_group}
\end{figure}

\begin{figure}[h]
	\begin{center}
		\includegraphics[width=1.0\textwidth]{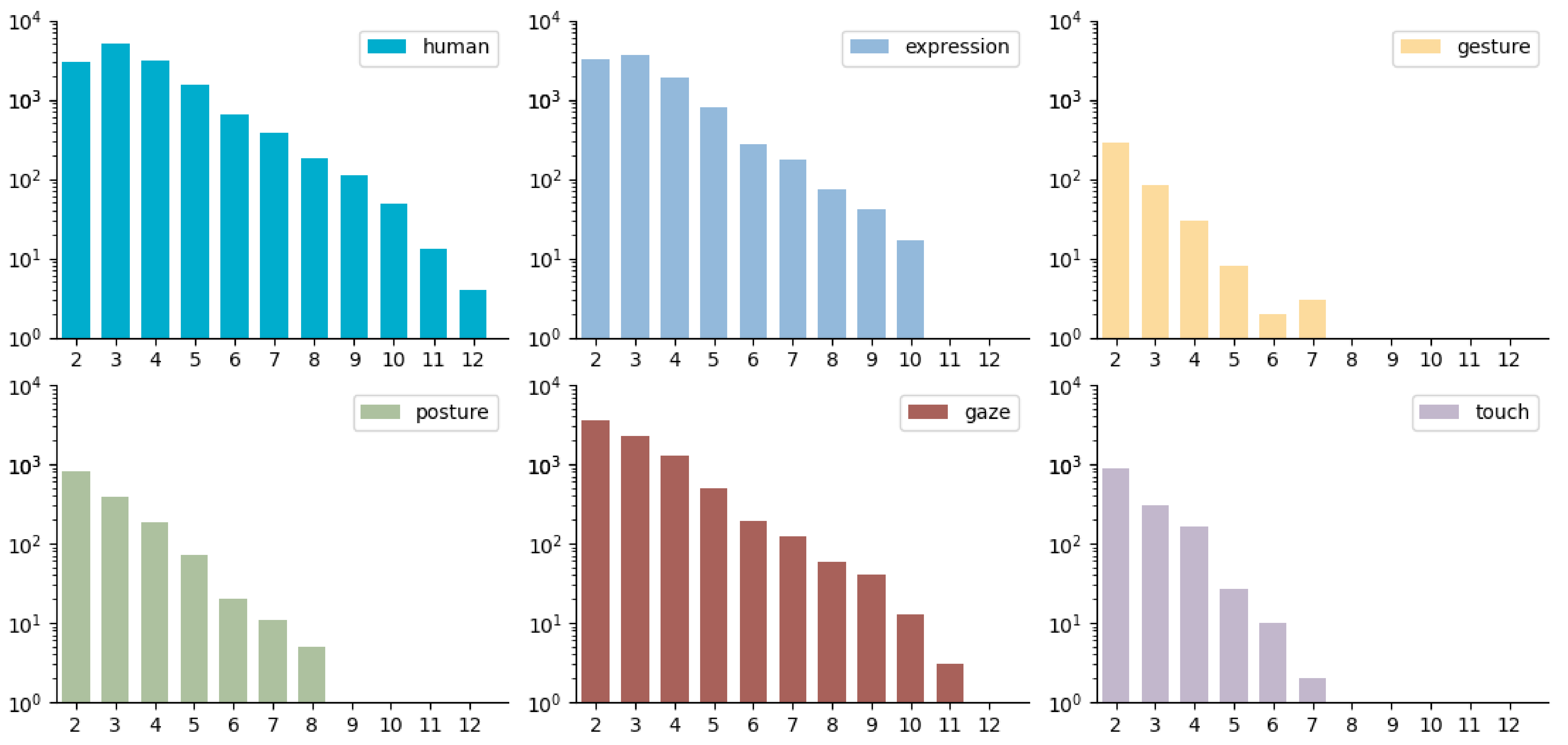}
	\end{center}
	\captionsetup{font=small}
	\caption{\small\textbf{The quantitative statistics} of human instances, gaze, touch, facial expression, gesture, posture in each image (\S\ref{sec:stat}).}
	\label{fig:dis_label}
\end{figure}

\noindent\textbf{More Statistics.}
We investigate the distribution of individuals engaged in group-wise interaction as illustrated in Fig.~\ref{fig:dis_group}.
It can be observed that the size of the gaze group exhibits considerable diversity, ranging from 2 to 12, while the touch group predominantly comprises 2 or 3 individuals. Furthermore, we present a detailed quantitative analysis of human behaviors depicted in each image (as shown in Fig.~\ref{fig:dis_label}), including the quantitative statistics of human instances, gaze, touch, facial expression, gesture, posture.

\begin{figure*}[t]
	\begin{center}
		\includegraphics[width=1.0\textwidth]{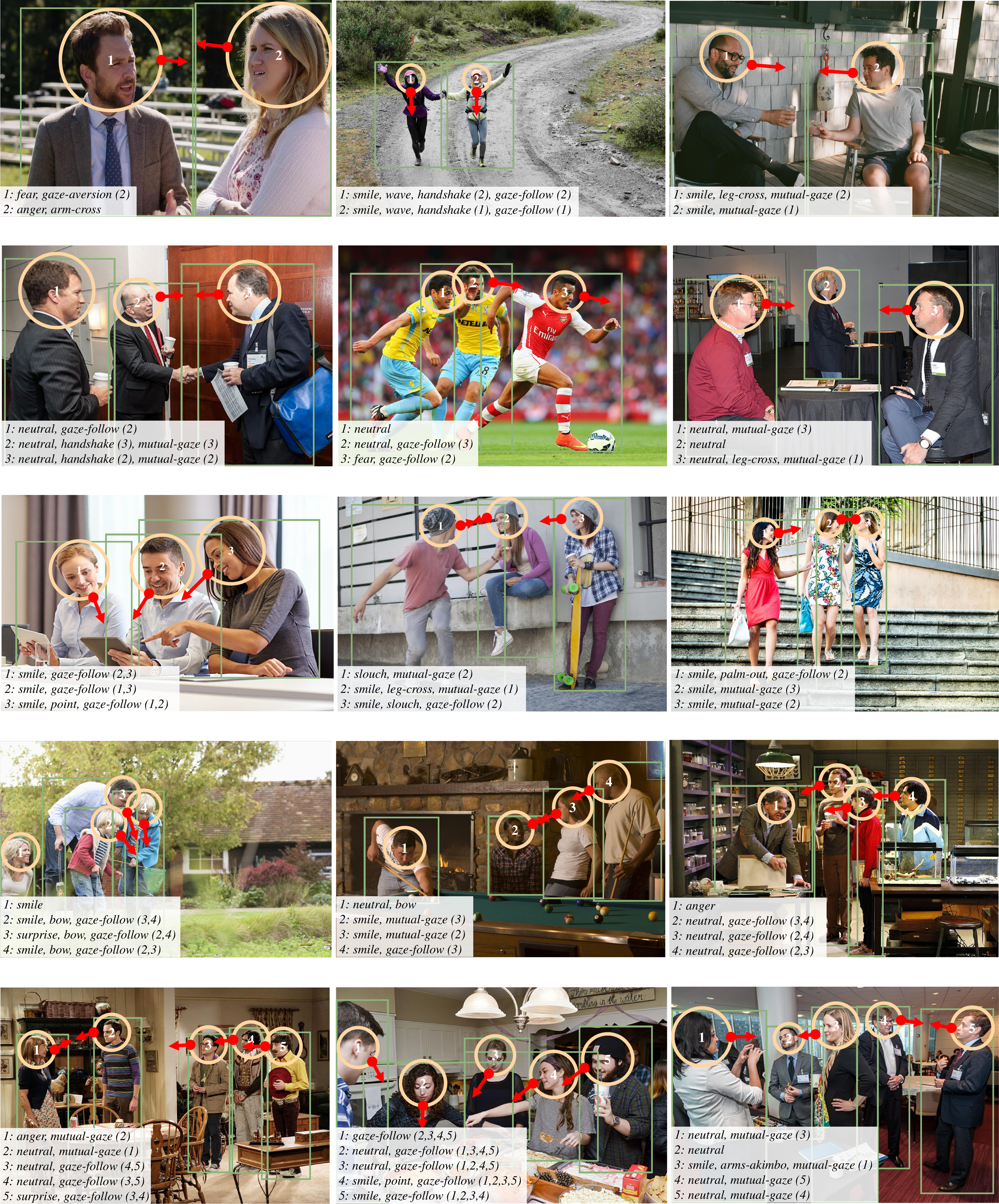}
	\end{center}
	\captionsetup{font=small}
	\caption{\small\textbf{Illustrative examples} of \dataset (\S\ref{sec:stat}).}
	\label{fig:examples}
\end{figure*}

\noindent\textbf{More Examples.}
To better showcase the diversity of our \dataset, we provide additional examples from various social environments involving diverse nonverbal interactions and different numbers of individuals, depicted in Fig.~\ref{fig:examples}.

\section{More Implementation Details on HOI-DET} \label{sec:hoi}
\noindent\textbf{Training Objective.} 
Following~\cite{tamura2021qpic,zhang2021mining,liao2022gen,ning2023hoiclip}, the HOI detection loss used in this work comprises four parts: a box regression loss $\mathcal{L}_b$, a generalized IoU loss $\mathcal{L}_\text{GIoU}$, a cross-entropy loss $\mathcal{L}_c^o$ for object classification and a cross-entropy loss $\mathcal{L}_c^a$ for action recognition.
The overall loss is the weighted sum of these parts: 
\begin{equation}\small\label{eq:interactiondecoder}
	\mathcal{L} = \lambda_{b} \mathcal{L}_\text{b} + \lambda_\text{GIoU} \mathcal{L}_\text{GIoU} + \lambda_c^o \mathcal{L}_c^o + \lambda_c^a \mathcal{L}_c^a,
\end{equation}
where $\lambda_{b}=2.5$, $\lambda_\text{GIoU}=1$, $\lambda_c^o=1$, $\lambda_c^a=2$.

\noindent\textbf{Evaluation Metrics.}
We adopt the mean Average Precision (mAP) for evaluation.
A HOI detection is considered a true positive when the human is correctly linked to the corresponding object using the appropriate verb, and this human-object pair is accurately localized (The accuracy of localization is evaluated by measuring the overlap between the bounding boxes).
For V-COCO \cite{gupta2015visual}, we report the mAPs for two scenarios: scenario 1 (S1) including the 4 body motions and scenario 2 (S2) excluding the HOI classes without object.
Regarding HICO-DET \cite{chao2018learning}, we assess performance across three settings: the complete set of 600 HOI categories (Full), a subset of 138 rare categories with fewer than 10 training images (Rare), and the remaining 462 categories (Non-rare).

\section{Additional Quantitative Results on \task} \label{sec:quant}
As seen in Table. \ref{tab:ic}, we conduct further analysis breaking down performance by interaction category. 
It can be observed that our \model demonstrates superior performance in all categories except the \textit{posture} category, with the marginal additional costs of our model.
It's worth noting that all models encounter a sharp performance decline for the \textit{gesture} category, which could potentially be attributed to the severe long-tailed  distribution within this category; for instance, \texttt{palm-out} and \texttt{beckon} are the two least frequent behaviors in \dataset.

\begin{table}[t]
	\caption{\small\textbf{Performance of interaction category} on \dataset \texttt{val} (\S\ref{sec:quant}).}
	\resizebox{\columnwidth}{!}{
			\tablestyle{3pt}{1.5}
			\begin{tabular}{c||c c|c c c c c}\hline\thickhline
				\rowcolor{mygray}
				Method & Params& FLOPs& Expression& Posture& Gesture& Touch& Gaze\\
				\hline
				m-QPIC& 41.42M& 56.10& 77.25& 66.91& 40.26& 80.68& 74.41\\
				m-CDN& 41.41M& 51.66& 77.98& 68.11& 39.25& 80.88& 73.52\\
				m-GEN-VLKT& 41.71M& 55.18& 78.91& \textbf{78.91}& 36.69& 81.37& 74.91\\
				NVI-DEHR(Ours)& 42.71M& 59.89& \textbf{79.37}& 72.94& \textbf{42.13}& \textbf{81.60}& \textbf{79.24}\\
				\hline
			\end{tabular}
		}
	\captionsetup{font=small}
	\label{tab:ic}
\end{table}

\section{Discussion} \label{sec:disc}
\noindent\textbf{Social Impact.}
\task takes a significant step towards creating socially-aware AI models with capabilities of generic nonverbal interaction understanding, and can benefit a variety of applications, like robotics, healthcare, and digital human.
The proposed \model and \dataset have no evident negative impact to society.
Nevertheless, there is a risk that someone could use it for malicious purposes, \eg, widespread surveillance, invasion of privacy, and potential abuse of personal information.
Therefore, we strongly advocate for the well-intended application of the proposed method, while simultaneously underscoring the importance of employing the dataset in a responsible and ethical manner.

\noindent\textbf{Limitation.} 
From a feasibility perspective, we carefully select the five most representative types and 22 subcategories of them to construct \dataset.
But, the constrained samples may fall short of capturing the full spectrum of nonverbal interactions that take place in real-world scenarios, which could hinder the applications of \task in more complex and diverse situations. 
Although our image-only \dataset, as a pionerring endeavor, is capable of delivering ample clues for the identification of nonverbal behaviors in most instances, there are occasional occurrences of ambiguity, like subtle facial expression and slight gaze-shift movements, akin to ambiguous actions like ``throw/catch frisbee'' in V-COCO. 

\noindent\textbf{Future Work.} 
Moving forward, we plan to extend our \dataset with temporal data for an in-depth analysis of nonverbal behaviors and enrich the variety of nonverbal interactions, like proximity \ie, the physical distances involved during the interactions \cite{hans2015kinesics}.
Inspired by previous works \cite{liu2020amplifying,zhang2022exploring,liu2022highlighting} in HOI-DET, which integrate simultaneous cues such as human pose or spatial relation from static images to mitigate label ambiguity, we intend to further exploit the co-occurrence of social signals in \task to recognize nonverbal interactions effectively.

\end{document}